\documentclass[letterpaper, 10pt, journal, twoside]{IEEEtran}  

\IEEEoverridecommandlockouts                              

\usepackage{amsmath} 
\usepackage{graphicx}
\usepackage{caption}
\usepackage{booktabs}
\usepackage{amssymb}
\usepackage{bm}

\usepackage[colorlinks,linkcolor=black, citecolor=black, urlcolor=black]{hyperref}
\usepackage{algorithmicx}
\usepackage[noend]{algpseudocode}
\usepackage{algorithm}
\usepackage{cite}
\usepackage{subfig}
\usepackage{url}
\usepackage[nameinlink,capitalise]{cleveref}
\usepackage{multirow}
\usepackage{xcolor}

\usepackage{balance}

\title{
	Efficient Trajectory Planning for Multiple Non-holonomic Mobile Robots 
	via Prioritized Trajectory Optimization
}

\author{Juncheng Li, Maopeng Ran, and Lihua Xie, \IEEEmembership{Fellow, IEEE}
	\thanks{Manuscript received: July, 4, 2020; Revised September, 7, 2020; Accepted November, 23, 2020.}
	\thanks{This paper was recommended for publication by Editor Nak Young Chong upon evaluation of the Associate Editor and Reviewers' comments. This work was supported by Delta-NTU Corporate Laboratory under the National Research Foundation Corporate Laboratory@University Scheme.}
	\thanks{Juncheng Li, Maopeng Ran, and Lihua Xie are with School of Electrical and Electronic Engineering, Nanyang Technological University, Singapore (email: juncheng001@ntu.edu.sg; mpran@ntu.edu.sg; elhxie@ntu.edu.sg). }%
	\thanks{Digital Object Identifier (DOI): see top of this page.}  
}

\begin{document}
	
	\markboth{IEEE Robotics and Automation Letters. Preprint Version. November, 2020} {Li \MakeLowercase{\textit{et al.}}: Efficient trajectory planning for multiple non-holonomic mobile robots}

	\makeatletter
	\let\ref\@refstar
	\makeatother
	
	\maketitle
	
	\begin{abstract}   
		
		In this paper we present a novel approach to efficiently generate collision-free optimal trajectories for multiple non-holonomic mobile robots in obstacle-rich environments. Our approach first employs a graph-based multi-agent path planner to find an initial discrete solution, and then refines this solution into smooth trajectories using nonlinear optimization. We divide the robot team into small groups and propose a prioritized trajectory optimization method to improve the scalability of the algorithm. Infeasible sub-problems may arise in some scenarios because of the decoupled optimization framework. To handle this problem, a novel grouping and priority assignment strategy is developed to increase the probability of finding feasible trajectories. Compared to the coupled trajectory optimization, the proposed approach reduces the computation time considerably with a small impact on the optimality of the plans. Simulations and hardware experiments verified the effectiveness and superiority of the proposed approach.
	\end{abstract}
	
	\section{Introduction}
	\IEEEPARstart{A}{utonomous} multi-robot systems are attracting significant attention from the industry since they can provide more diverse functionality and efficiency than single-robot systems. Many applications of multi-robots systems require coordination of mobile robots navigating in a complex environment, e.g., material handling in warehouses \cite{guizzo2008three} and drone delivery \cite{dorling2016vehicle}. In these scenarios, collision-free trajectories connecting initial and final positions for the robots need to be generated, which is referred to as the labeled multi-robot trajectory planning problem \cite{tang2016safe}.
	
	It is challenging to obtain optimal trajectories for a large team of robots in an efficient way. {In \cite{mellinger2012mixed,augugliaro2012generation}, the optimal trajectory generation problem is formulated as mixed-integer quadratic programming (MIQP) or sequential convex programming (SQP) problems. These methods try to jointly optimize the trajectories of all robots.} Though the optimality is guaranteed, the application of these methods is limited to small teams with few obstacles in the environment.

	To improve computational efficiency, decoupled planning methods have been proposed. A widely used decoupled scheme for multi-robot trajectory planning is {sequential planning}\cite{vcap2015prioritized}. {In sequential planning, the trajectory of each robot is decoupled and coordinated by avoiding the previously planned robots. The trajectory optimization methods in \cite{chen2015decoupled,robinson2018efficient} utilize sequential planning to decouple inter-robot collision constraints and achieve significant improvement in computational efficiency.} These decoupled planning methods are relatively fast but typically suffer from incompleteness. In certain scenarios, a feasible solution for the robots exists but cannot be found.
	
	In obstacle-rich environments, the trajectory planning problem is generally solved using a two-stage pipeline\cite{ding2019efficient}, i.e., path finding and trajectory optimization. A great number of works have been done in single-robot cases\cite{liu2017planning,gao2018online,zhou2019robust}. The basic idea of such methods is to first generate a geometric path and then optimize the path to a smooth and dynamically feasible trajectory. This two-stage pipeline can also be applied in the multi-robot trajectory planning problem. In \cite{tang2018hold,honig2018trajectory, park2020efficient}, collision-free discrete paths are first generated for all robots. Then based on the results, a quadratic program (QP) problem is formulated for each robot to generate the optimal trajectory. The multi-robot trajectory planning algorithms using the two-stage pipeline are guaranteed to be complete \cite{tang2018hold,honig2018trajectory}. Besides, since the path finding stage provides a good initial guess for the multi-robot trajectory optimization, the efficiency of solving the optimization problem is significantly improved by using the two-stage pipeline.

	The aforementioned multi-robot trajectory planning algorithms \cite{tang2018hold,honig2018trajectory, park2020efficient} focus on robots with linear dynamics. In this case, the trajectory optimization can be formulated as a QP problem, which is convex and easy to solve. However, in modern industry, most mobile robots are differential-drive and inherently subject to nonlinear dynamics. When nonlinear dynamics is considered, the trajectory optimization can only be formulated as a general nonconvex nonlinear programming (NLP) problem, which makes the existing multi-robot trajectory planning methods \cite{tang2018hold,honig2018trajectory, park2020efficient} inapplicable.  Currently, the problem of motion planning for multiple differential-drive robots is generally addressed via discrete formulations\cite{yu2018effective,honig2019persistent,motes2020multi}. However, since the planned piecewise linear paths contain corner turns that are dynamically infeasible for differential-drive robots, these paths are difficult for the robots to execute.

	The multi-robot motion coordination problem can also be solved in a distributed way. Reactive methods such as buffered Voronoi cells\cite{zhou2017fast} and velocity obstacle (VO)\cite{van2008reciprocal,van2011reciprocal} are developed. Besides, in \cite{luis2019trajectory,luis2020online}, distributed trajectory planning methods are proposed based on the distributed model predictive control (DMPC). In these methods, the robot follows the shortest path to its destination and resolves conflicts in real time when future collisions are detected. Some algorithms have been extended to nonlinear dynamics and applied to non-holonomic robots\cite{bareiss2015generalized,alonso2018cooperative}. Distributed planning methods are computational efficient, but cannot guarantee no deadlock and are poorly suited to problems in maze-like environments.

	Based on the above considerations, in this paper, we aim to propose an efficient trajectory planning approach that generates safe, dynamically feasible and near-optimal trajectories for multiple non-holonomic mobile robots. Our approach first uses a multi-robot path planner to generate an initial solution for the problem, and then it is refined into smooth trajectories by solving a trajectory optimization problem. In both stages, the nonlinear motion model of the mobile robots is directly considered to guarantee the feasibility of the trajectories. In particular, we introduce a prioritized trajectory optimization method to improve the computational efficiency such that the algorithm is applicable to large-scale robot teams. To the best of our knowledge, this paper is the first attempt to develop a centralized multi-robot trajectory planning method for non-holonomic mobile robots in continuous space. The main contributions of this work are as follows:
	
	\begin{itemize}
		\item{An efficient multi-robot trajectory planning approach which generates collision-free optimal trajectories for a large team of non-holonomic mobile robots.
		}
		\item{A prioritized optimization method which decouples the multi-robot trajectory optimization problem and improves the computational efficiency significantly.
		}
		\item{Extensive evaluations of the proposed approach via simulations and real-world experiments.
		}
	\end{itemize}
	
	The remaining of this paper is organized as follows. In Section \ref{problem_formulation}, we state the problem formulation. In Section \ref{approach}, the proposed approach is described in detail. We evaluate our approach in Section \ref{simulation} and present a real-world experiment in Section \ref{experiment}. Finally, Section \ref{conclusion} concludes the paper and discusses future work.

	\section{Problem Formulation}\label{problem_formulation}
	
	Consider a multi-robot system consisting of $N$ mobile robots which operate in a 2-D workspace $\mathcal{W} \subseteq \mathbb{R}^2$. The obstacles in the environment are assumed to be known and denoted as $\mathcal{O}$. The free workspace of the robots is given by $\mathcal{F} = \mathcal{W} \backslash \mathcal{O}$. The collision model of each robot is defined as a circle with radius $R$. The subset of $\mathcal{W}$ occupied by the body of a robot at position $x \in \mathbb{R}^2$ is denoted by $\mathcal{R}(x)$. For each robot $i$, a task is assigned to move from its start position $s_i \in \mathcal{F}$ to its goal position $g_i \in \mathcal{F}$. To guarantee that there is no inevitable collision at the start and goal positions, we assume the tasks ${\left\langle s_i,g_i\right\rangle}_{i=1\ldots N}$ must satisfy $\mathcal{R}(s_j) \cap \mathcal{R}(s_k)=\emptyset$ and $\mathcal{R}(g_j) \cap \mathcal{R}(g_k)=\emptyset$ for all $j \ne k$.
	
	A sequence of waypoints which connect the start and goal positions of the robot is denoted by path $p={\{r^k\}}_0^{N_p}$, where $r^k=[x^k,y^k,\theta^k]^\text{T}$ denotes the $k$th waypoint on the path and $N_p$ denotes the total number of waypoints. A trajectory $\nu$ is path $p$ parameterized by time $t$. Let $\text{pos}(r)$ denote the 2-D position of waypoint $r$. For each robot $i$, its body should not collide with any obstacle in the environment when following its trajectory $\nu_i$, i.e., $\mathcal{R}(\text{pos}(r_i(t))) \subseteq \mathcal{F}$. Besides, the collisions between any two robots should be avoided, i.e., $\mathcal{R}(\text{pos}(r_i(t))) \cap \mathcal{R}(\text{pos}(r_j(t)))=\emptyset, \ \forall i\ne j$. 
	
	Furthermore, the planned trajectory should be dynamically feasible for each robot. The kinematic model of each robot is defined as
	\begin{equation}\label{equ:input_constraint}
	\dot z=f(z,u), \ u \in \mathcal{U}.
	\end{equation}
	In this paper, the unicycle model for differential wheeled robots is considered. The state $z$ is defined as $[x,y,\theta]^\text{T}$ which consists of the 2-D position and orientation. The robot is controlled by the linear and angular velocity $u=[v,\omega]^\text{T}$, and its motion equations are given by
	\begin{equation}\label{equ:kinematics}
	\dot x=v\cos(\theta), \ \dot y=v\sin(\theta), \ \dot \theta=\omega.
	\end{equation}
	Besides, each robot is limited by its maximum velocity, i.e., $\mathcal{U}_i = \{ {[v_i,\omega_i]^{\textrm{T}}}: \left| v_i \right| < {v_i^{\max }},\left| \omega_i \right| < {\omega_i^{\max }}\}$. In this paper, we aim to generate dynamically feasible and optimal trajectories $\nu_1,\ldots,\nu_N$ for a group of mobile robots with nonlinear dynamics, such that each robot can reach its goal, and avoid collisions with obstacles and other robots.

	\section{Methodology}\label{approach}

	\begin{figure*}
		\centering
		\includegraphics[width=0.96\linewidth,trim={5 5 5 5},clip]{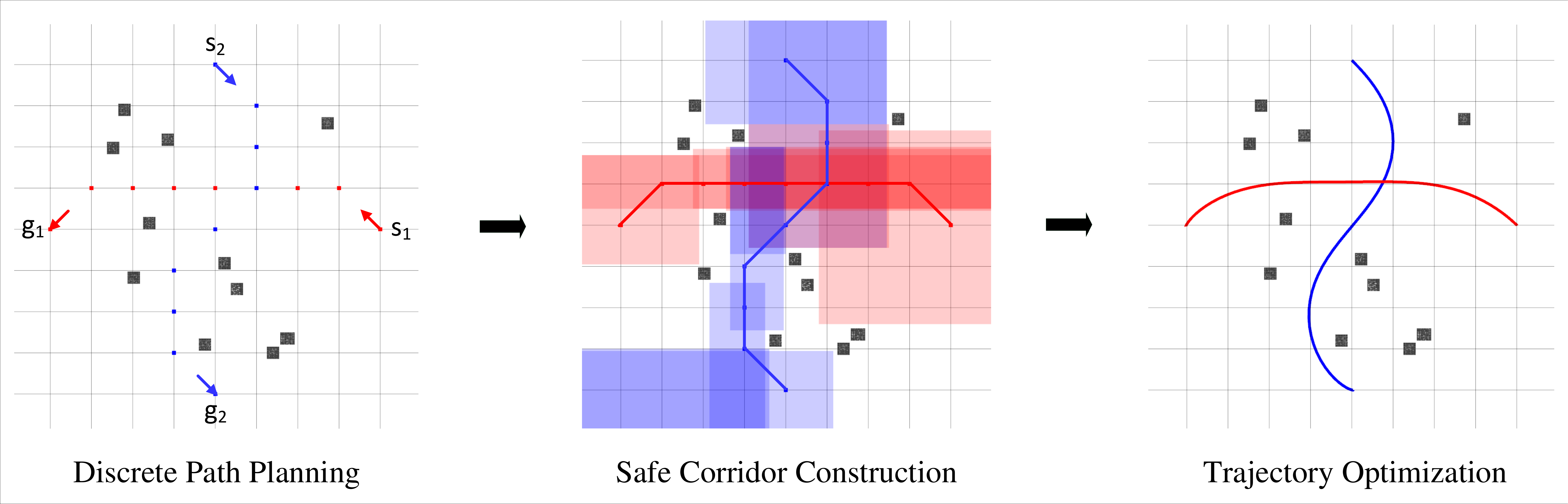}
		\caption{Overall architecture of the proposed multi-robot trajectory planning approach. Gray squares show the static obstacles in the environment. In this example, the start and goal positions of agent 1 and 2 are assigned as $s_1=(4,0),\, g_1=(-4,0),\,s_2=(0,4),\,g_2=(0,-4)$. Firstly, we find shortest collision-free paths for the robots. Secondly, safe corridors (red and blue blocks) are constructed along the planned paths. Finally, a constrained trajectory optimization problem is solved to generate smooth and dynamically feasible trajectories for the robots.
		}
		\label{fig:framework}
	\end{figure*}
	
	The overall architecture of the proposed multi-robot trajectory planning approach is shown in Fig. \ref{fig:framework}. Firstly, we use a graph-search based method to find shortest collision-free paths for all robots with the non-holonomic constraint \eqref{equ:kinematics}. Then the safe corridor is constructed around each robot's path, which is a collection of convex polyhedra that models the safe space of the robot. Finally, we formulate a constrained nonlinear optimization problem based on the planned discrete paths and safe corridors. By efficiently solving the problem using prioritized trajectory optimization, safe and near-optimal trajectories for all robots are obtained. The detail of our approach is described in the following subsections.

	\subsection{Discrete Path Planning}\label{sec:discrete}
	
\begin{figure}
		\vspace{-0.4cm}
		\centering
		\subfloat[]{
			\label{fig:4_connected}
			\includegraphics[width=0.33\linewidth,trim={10 5 10 5},clip]{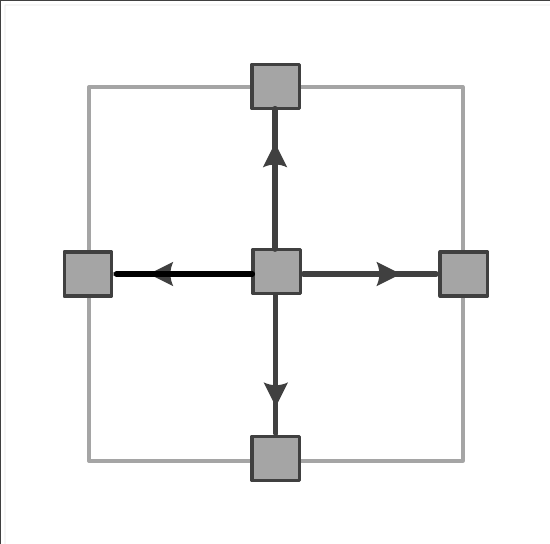}
		} \hspace{0.3cm}
		\subfloat[]{
			\label{fig:motion_primitives}
			\includegraphics[width=0.36\linewidth,trim={10 14 10 10},clip]{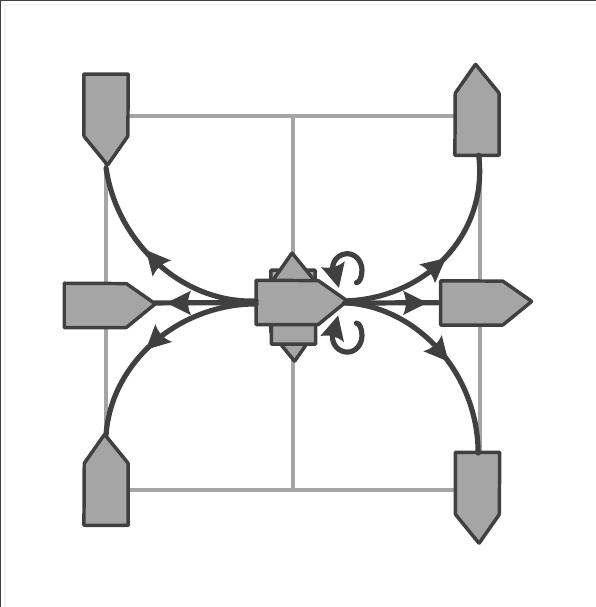}
		}
		\caption{Edges in the grid graph. Fig. 2a shows the traditional 4-connected edges. Considering the kinematic constraint of differential wheeled robots, the edges in the graph are defined by motion primitives, as shown in Fig. 2b.}
	\end{figure}

	The workspace of the robots is abstracted as an undirected graph $\mathcal{G}=\{\mathcal{V},\mathcal{E} \} $ where each vertex $v \in \mathcal{V}$ corresponds to a location agents can cover and each edge $(v_i,v_j) \in \mathcal{E}$ corresponds to a path the agents traverse when moving from vertex $v_i$ to $v_j$. In this work, occupancy map of the environment is transformed into a 2-D grid graph. In general, each vertex and one of its 4 neighbors can form an edge (Fig. \ref{fig:4_connected}). As mentioned before, the state of each robot contains both the position and orientation information. The traditional grid graph only considers the position and neglects the orientation information, so it is not suitable for non-holonomic mobile robots. To handle this problem, we propose a new graph representation for the multi-robot path planning.
	
	Firstly, in the graph $\mathcal{G}$, each vertex $v\in \mathcal{V}$ is now three dimensional and corresponds to a robot state consisting of the position and orientation. The heading of the robot can be chosen from the four main directions, i.e., north, south, east, and west. Let ${\text{P}(v)}$ denote the 2-D position of vertex $v$, then each vertex should satisfy $\mathcal{R}(\text{P}(v)) \subseteq \mathcal{F}$ in order to avoid possible collisions. Secondly, based on the kinematic model \eqref{equ:kinematics}, we define the possible one-step state transitions of the robot, which is also referred to as motion primitives\cite{pivtoraiko2011kinodynamic}. There are three available actions for each robot to take at each step, i.e., moving forward, moving backward and turning 90 degrees. Combining these possible actions, we can obtain all the motion primitives which are shown in Fig \ref{fig:motion_primitives}. Let $\mathcal{A}$ denote the set of all the motion primitives. Each one-step transition in $\mathcal{A}$ is considered as an edge in the graph $\mathcal{G}$. The edge is valid only when no collision is found when any one of the robots traverses that edge. 
	
	Since the kinematics of the robot is considered in the graph construction process, the feasibility of the motion primitives needs to be guaranteed. Let $D$ denote the grid size of graph $\mathcal{G}$, $\Delta T$ denote the time required for the robot to traverse one edge. Due to the physical limits of each robot, the following condition should be satisfied to guarantee the feasibility of the diagonal one step transition:
	\begin{equation}\label{equ:feasiblility}
	v^{\max }_i\Delta T \ge \frac{\pi }{2}D, \ \ {w^{\max }_i\Delta T \ge \frac{\pi }{2}}	
	, \ \ \forall i \in \{1,\cdots, N\}.
	\end{equation}
	It can be observed that if the diagonal one-step transition is guaranteed to be feasible, then all actions in $\mathcal{A}$ are feasible. Since both $\Delta T$ and $D$ in \eqref{equ:feasiblility} are user-defined parameters, we can construct suitable graph representations for different scales of environments based on the proposed method.

	Conflicts may occur when each robot plans its path individually. In the same graph $\mathcal{G}$ which is shared by all robots, two kinds of conflicts are considered, i.e., vertex conflict and edge conflict. At each timestep $k$, if two robots $i$ and $j$ occupy two vertexes which are too close to each other, it is a vertex conflict denoted by $\left\langle i,j,k \right\rangle$. At any time between two consecutive timesteps $k_1$ and $k_2$, if two robots are too close to each other when traversing their own edge, it is an edge conflict denoted by $\left\langle i,j,k_1,k_2 \right\rangle$. The vertex conflict set at timestep $k$ and edge conflict set at timestep $(k_1,k_2)$ are described as:
	\begin{subequations}\label{equ:vcon_econ}
		\begin{align}
		\text{V}&\text{Con}(k)=\{\left\langle i,j \right\rangle | \\ &\ \; \; \mathcal{R}(\text{pos}(r_i^k)) \cap  \mathcal{R}(\text{pos}(r_j^k)) \neq  \emptyset\},\nonumber\\
		\text{E}&\text{Con}(k_1,k_2)=\{\langle i,j \rangle | \\ &\mathcal{R}(\text{pos}(r_i(t))) \cap  \mathcal{R}(\text{pos}(r_j(t))) \neq \emptyset, \forall t \in (k_1 \Delta T,k_2 \Delta T) \}.\nonumber
		\end{align}
	\end{subequations}
	
	Finding shortest conflict-free paths for multiple agents in the defined graph is known as a multi-agent path finding (MAPF) problem. 
	Solving MAPF optimally is NP-hard \cite{honig2019persistent} and suffers from a scalability problem. In this work, enhanced conflict-based search (ECBS)\cite{barer2014suboptimal} is leveraged as the multi-robot discrete path planner. This algorithm can obtain a bounded suboptimal solution efficiently and guarantee completeness.
	
	ECBS works in two-levels. At the high-level, a binary constraint tree (CT) is constructed to resolve the detected conflicts. At the low-level, optimal paths for individual agents are planned which are consistent with their own constraints. In the beginning, the root node of the CT contains no constraints and the low-level search returns an initial solution. The solution is checked for conflicts in chronological order by \eqref{equ:vcon_econ}. Suppose a vertex conflict $\left\langle i,j,k\right\rangle$ is found, then two 
	child nodes are generated to resolve this conflict. The first node adds a constraint for agent $i$ to avoid staying at $r_i^k$ at timestep $k$. The second node adds a constraint for agent $j$ to avoid staying at $r_j^k$ at timestep $k$. An edge conflict $\left\langle i,j,k_1,k_2 \right\rangle$ can be resolved in a similar way. Each time a CT node is created, the low-level search for the agent with added constraint is executed, which returns a new solution. ECBS performs a best-first search on the CT where nodes are ordered by their costs. The expansion of the CT continues until a solution without any conflict is found. By implementing a focal search with conflict heuristic in both high-level and low-level, ECBS solves the MAPF problem efficiently.

	Since the relative distance between robots will be checked at each timestep in the trajectory optimization process, the length of each robot's discrete path should be the same. In this case, we denote the length of the longest path as $M$. For each agent $i$, we append its goal position $g_i$ at the end of its path $p_i$ until the length of $p_i$ is $M$.

	\subsection{Safe Corridor Construction}\label{sec_SFC}
	
	\begin{figure}
		\centering
		\includegraphics[width=0.6\linewidth,trim={0 57 0 23},clip]{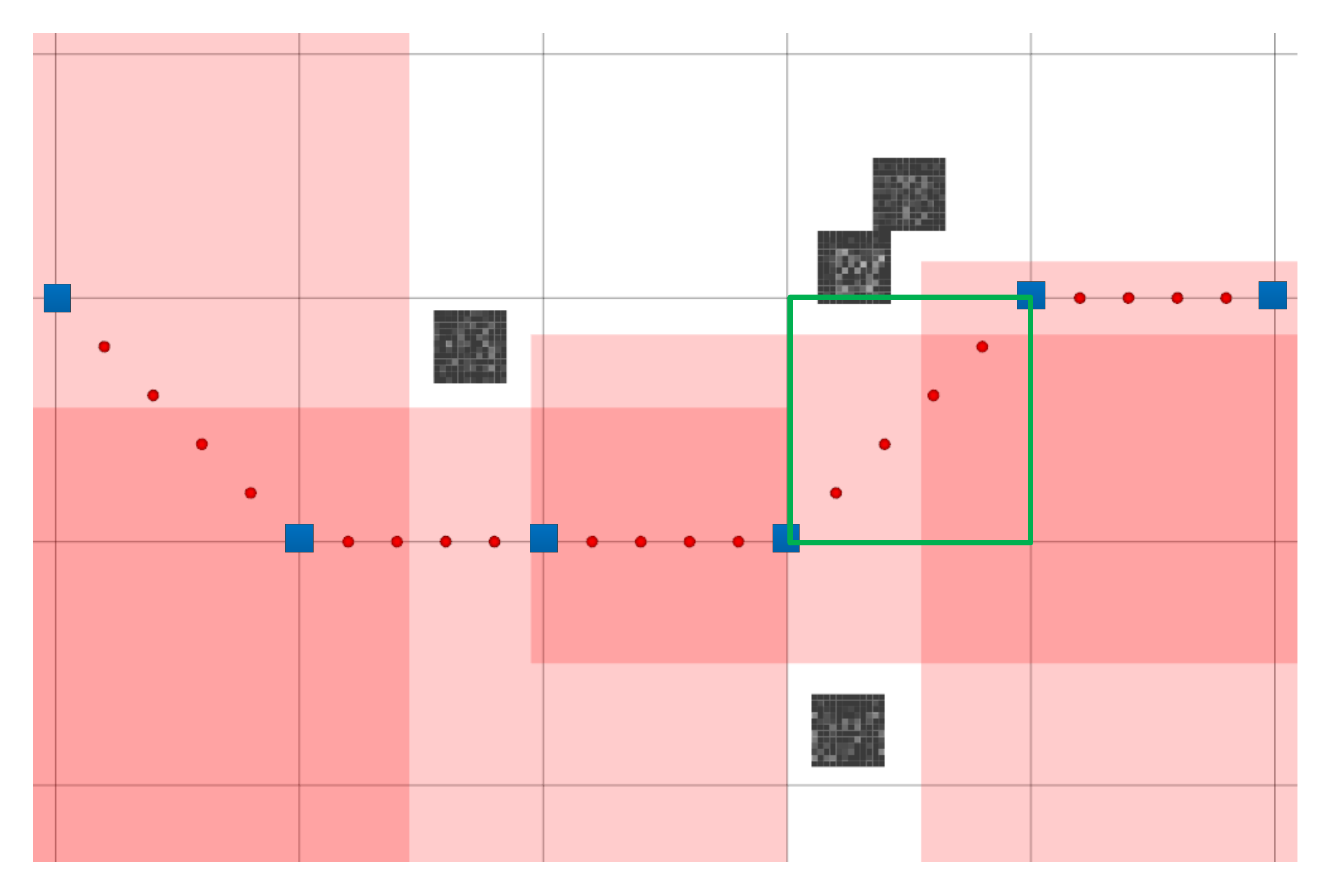}
		\caption{ Construction of the safe corridor. Blue squares show an original discrete path, which cannot be directly used to construct the safe corridor. The green box shows a failure case. Red dots show a sampled path. By expanding each point of this path in $x$ and $y$ axes, the safe corridor can be constructed successfully, which is shown as red blocks.}
		\label{fig:SFC}
	\end{figure}
	
	The planned discrete path of each robot is denoted by $p=\{r^k\}^{M}_0$, where each waypoint $r^k$ is in the free space $\mathcal{F}$. The $k\text{th}$ line segment in the path is denoted by $I^k=\langle r^{k-1} \rightarrow r^{k}\rangle$. We generate a convex polyhedron around each line segment in the path to construct a valid safe corridor. The generated convex polyhedron around the line segment $I^k$ is denoted as $\mathcal{S}^k$. To ensure an agent in the polyhedron $\mathcal{S}^k$ does not collide with any obstacle in the environment, the following condition should be satisfied:
	\begin{equation}\label{equ:SFC}
	\mathcal{R}(a) \cap \mathcal{O} = \emptyset, \ \forall a \in \mathcal{S}^k.
	\end{equation}
	The collection of these convex polyhedra constitutes the safe corridor, which is denoted by $\text{SC}(p)=\{\mathcal{S}^k \,|\, k=1, \dots M \}$. Note that the safe corridor needs to be sequentially connected, and hence satisfies the following condition:
	\begin{equation}
	\mathcal{S}^k \cap \mathcal{S}^{k+1} \neq \emptyset, \ \forall k \in \{1,\cdots,M-1\}.
	\end{equation}
	
	\begin{algorithm}
		\caption{Safe corridor construction}
		\label{alg:sfc}
		\begin{algorithmic}[1] 
			\Require discrete path $p=\{r^k\}_0^H$
			\Function {Safe corridor }{$p$}
			\For {$k \gets 1$ to $H$}
			\If {$r^k \in \mathcal{S}^{k-1}$}
			\State $\mathcal{S}^k \gets \mathcal{S}^{k-1}$
			\Else
			\State $\mathcal{S}^k \gets r^k$
			\State $\mathcal{D} \gets \{+x, -x, +y, -y\}$
			\While {$\mathcal{D} \neq \emptyset$}
			\For {$d$ in $\mathcal{D}$}
			\State $\widetilde{\mathcal{S}}^k \gets$ expand $\mathcal{S}^k$ in the direction $d$ 
			\If{$\mathcal{S}^k \subseteq \mathcal{F} $}
			\State $\mathcal{S}^k \gets$$\widetilde{\mathcal{S}}^k$
			\Else
			\State $\mathcal{D} \gets \mathcal{D} \backslash d$
			\EndIf
			\EndFor
			\EndWhile
			\EndIf
			\EndFor
			\State \Return{$\text{SC}(p)=\{\mathcal{S}^k | k=1,\cdots,H\}$}
			\EndFunction
		\end{algorithmic}
	\end{algorithm}
	
	An axis-search method inspired from \cite{park2020efficient} is leveraged in this work to build the safe corridor. The safe corridors are expanded in both $x$ and $y$ axes until the maximum possible free space has been covered. To ensure the continuity of the safe corridors, $\mathcal{S}^k$ should be initialized to be the set containing the consecutive waypoints $r^{k-1}$ and $r^k$. In our approach, since the motion primitives are included in the discrete path planning stage, the planned paths consist of horizontal, vertical and diagonal line segments.  However, the initial safe corridor around a diagonal line segment may not be completely within the free space. Fig. \ref{fig:SFC} shows an example.

	To solve this problem, firstly we divide each line segment in the path into $h$ equal parts. The new sampled path now consists of $H=1+h(M-1)$ waypoints. Then we initialize $S^k$ to be the point $r^k$, and expand it in $x$ and $y$ axes until the maximum possible free space is covered. The safe corridor initialized in this way is completely within the free space. To guarantee the continuity of the safe corridor, the minimum safe regions around two consecutive waypoints need to be overlapped, which means that the following condition should be satisfied:
		\begin{equation}
		\frac{\sqrt 2D}{h} <  2R_i,\ \ \forall i \in \{1,\cdots, N\},
		\end{equation}
	where $R_i$ is the radius of the collision model of the robot $i$. Additionally, if the point $r^{k}$ is found to be within the previous safe region $\mathcal{S}^{k-1}$, the safe corridor expansion process is not needed and the convex polygon $\mathcal{S}^{k}$ should be exactly the same as $\mathcal{S}^{k-1}$. The whole safe corridor construction process of each robot's path is summarized in Alg. \ref{alg:sfc}.

	\subsection{Trajectory Optimization Problem}
	
	The discrete paths are refined into smooth and feasible trajectories in the trajectory optimization phase. For each robot, we assign a time $t^k=k{\Delta t}$ to each waypoint in its sampled path $p=\{r^k\}_0^H$ so that the path becomes a reference trajectory $\nu^r=\{r^k,t^k\}_0^H$. As mentioned before, each line segment in the original path is divided into $h$ equal parts, so $\Delta t=\Delta T /h$. Based on the reference trajectory $\nu^r$ and the constructed safe corridor $\text{SC}(p)$, we aim to obtain the optimal trajectory $\nu=\{z^k,t^k\}_0^H$ for each robot such that the group of robots can reach their goals and avoid any collision. The nonlinear optimization problem is formulated as:
	\begin{subequations}\label{equ:MPC_new}
		\begin{align}
		\text{minimi}& \text{ze} \label{equ:MPC_cost} \ \sum\limits_{i = 1}^N ( \sum\limits_{j = 1}^{H-1} \Delta {u_i^j}^{\mathrm{T}}P\Delta u_i^j+ \sum\limits_{j = 1}^H {{{}\widehat{z}_i^j}^{\mathrm{T}}Q{}\widehat{z}_i^j)} 
		\\
		\text{s.t.} \; \; \; &{z_i^0} = s_i, \; z_i^H=g_i \label{equ:MPC_inital} \qquad  \quad  \; \ \,  \forall i \\
		&{z^{j + 1}_i} = f({z^j_i},{u^j_i}), 
		\label{equ:MPC_state_transition} \quad \qquad  \;   \; \forall i,j \\
		&\text{pos}({z_i^j}) \label{equ:state_constraint} \in \mathcal{S}_i^j, \qquad \qquad \  \ \, \, \, \forall i,j\\
		&{u_i^j} \label{equ:MPC_input_constraint} \in \mathcal{U}_i,\qquad \qquad \qquad  \
		\ \ \, \,    \forall i,j \\
		&\hspace{-0.1cm}\left\| {\text{pos}(z_{i_1}^j)-\text{pos}(z_{i_2}^j)} \label{equ:relative_constraint} \right\| \ge R_{i_1}+R_{i_2}, \,  \forall i_1,i_2,j,i_2\neq i_1
		\end{align}
	\end{subequations}
	where $\Delta u_i^j=u_i^j-u_i^{j-1}$, $\widehat{z}_i^j=z_i^j-r_i^j$, and $P\in \mathbb{R}^{2\times 2}$ and $Q \in \mathbb{R}^{3\times 3}$ are two positive definite weighting matrices. The objective function consists of two parts. In the first part, differences between two successive control inputs are penalized for the smoothness of the trajectories. In the second part, since the reference trajectory $\nu^r$ of each robot is already a feasible solution, to keep the feasibility of the nonlinear optimization problem \eqref{equ:MPC_new}, we penalize the deviation between the optimal trajectory and the reference trajectory. The state transition at each timestep is determined by \eqref{equ:MPC_state_transition}, where the nonlinear model $f$ is the discrete-time version of \eqref{equ:kinematics}. The position of each waypoint on the trajectory is limited by the constructed safe corridor in the constraint \eqref{equ:state_constraint}. The control inputs of each robot is limited to physically admissible values in \eqref{equ:MPC_input_constraint}. Moreover, the safe distance between each pair of robots is limited by \eqref{equ:relative_constraint}.

	\subsection{Prioritized Trajectory Optimization}

	Problem \eqref{equ:MPC_new} is not efficiently solvable for large-scale robot teams. Therefore, we propose an efficient prioritized trajectory optimization method to solve the problem. The robots are first divided into some groups with unique priorities. Then the trajectory optimization subproblem is solved sequentially from the highest-priority group to the lowest-priority group. In each iteration, trajectories of the current group of robots are optimized under the constraint that they must avoid collisions with all the higher-priority robots. The trajectory optimization process runs relatively fast using this decoupled framework. However, if the priority is not carefully defined based on the specific scenarios, prioritized optimization may lead to infeasible subproblems for lower-priority robots due to lack of consideration by higher-priority robots. To handle this problem, a novel grouping and priority assignment strategy is proposed to enable the algorithm to find a near-optimal solution with a higher probability.
	
	\begin{algorithm}
		\caption{Grouping and priority assignment}
		\label{alg:sequential_optimization}
		\begin{algorithmic}[1] 
			\Require reference trajectories $\nu^r_1\ldots\nu^r_N$
			\Function {Priority assignment}{$\{\nu^r_i\}_{i=1\ldots N}$}
			\For {$t \gets 1$ to $H$}
			\State $(i_1,i_2,i_3)^t \gets$ find triple satisfies \eqref{three_conflict}
			\State Insert $(i_1,i_2,i_3)$ into $L$
			\EndFor
			\While {$L \neq \emptyset$}
			\State $e \gets$ most common element in $L$
			\State Append $e$ to the group list $G$
			\For {$l$ in $L$}
			\For {robot $m$ in $e$}
			\If{robot $m$ in $l$}
			\State Remove $m$ from $l$ 
			\EndIf
			\EndFor
			\EndFor
			\EndWhile
			\For {$n \gets 1$ to $N$}
			\If {$n$ not in $G$}
			\State Append $n$ to $G$
			\EndIf
			\EndFor
			\State \Return{$G$}
			\EndFunction
		\end{algorithmic}
	\end{algorithm}

	The main concern of the prioritized trajectory optimization is the inter-robot constraint \eqref{equ:relative_constraint}. Since the optimal trajectories are close to the reference ones according to the cost function \eqref{equ:MPC_cost}, we can analyze the inter-robot constraints based on the reference trajectory $\nu^r_1,\ldots,\nu^r_N$. At each timestep $t$, we search for a triple of robots $(i_1,i_2,i_3)^t$ which satisfies:
	\begin{equation}\label{three_conflict}
		\left\| \text{pos}(r_a^t) - \text{pos}(r_b^t) \right\| \le D_{\text{th}}, \; \forall a,b\in \{i_1,i_2,i_3\},b \neq a
	\end{equation}
	where $D_{\text{th}}=\sqrt 2 D$ is a threshold. Each triple represents a situation where three robots are close to each other at a specific timestep. In the discrete path planning stage, only when four robots are located at the four vertexes of a grid cell at the same time, we can find a situation where four agents simultaneously satisfy \eqref{three_conflict}. Therefore, in this case, we only analyze inter-robot collisions among three robots rather than four or more robots.

	If the robots in a triple are assigned different priorities, the higher priority robots will plan their trajectories first. Then these trajectories are considered as hard constraint in the optimization problem of the low priority robots. In this case, the feasibility of the original problem may be lost. Therefore, the three robots in a triple should form a group and be assigned the same priority. The list of all triples satisfying $\eqref{three_conflict}$ is denoted by $L$. 
	
	If the number of robots in the workspace is large, $L$ will contain a great number of elements and thus a robot may be included in different triples. In this case, the robots cannot be grouped directly. Since each triple represents a situation where three robots are close to each other, if one triple repeats many times in $L$, that means the trajectories of these three robots are highly coupled. A group of robots with higher coupling effect should be assigned with a higher priority, so we propose a priority assignment algorithm based on the number of occurrences of each triple.

	In the beginning, the most common element in $L$ is selected as the first-priority group. Since the robots in this new group may be included in other triples in $L$, we remove all of them from $L$ and obtain a new list $L'$. As a result, each element of $L'$ may contain three, two, or just one robot. Then we continue to select the most common element in the new list $L'$ as the second-priority group. The process is repeated until the list is empty. Finally, we need to check the completeness of the algorithm. If a robot does not have any significant coupling with other robots, then it will not be included in the original list $L$. In this case, the robot is regarded as a single-robot group and assigned the lowest priority. Using the proposed algorithm, all robots in a team can be grouped and assigned priorities successfully and completely. The whole process is summarized in Alg. \ref{alg:sequential_optimization}.

	\begin{figure}
		\vspace{-0.3cm}
		\centering
		\subfloat[]{
			\label{fig:sequential_1}
			\includegraphics[width=0.39\linewidth,trim={10 5 5 5},clip]{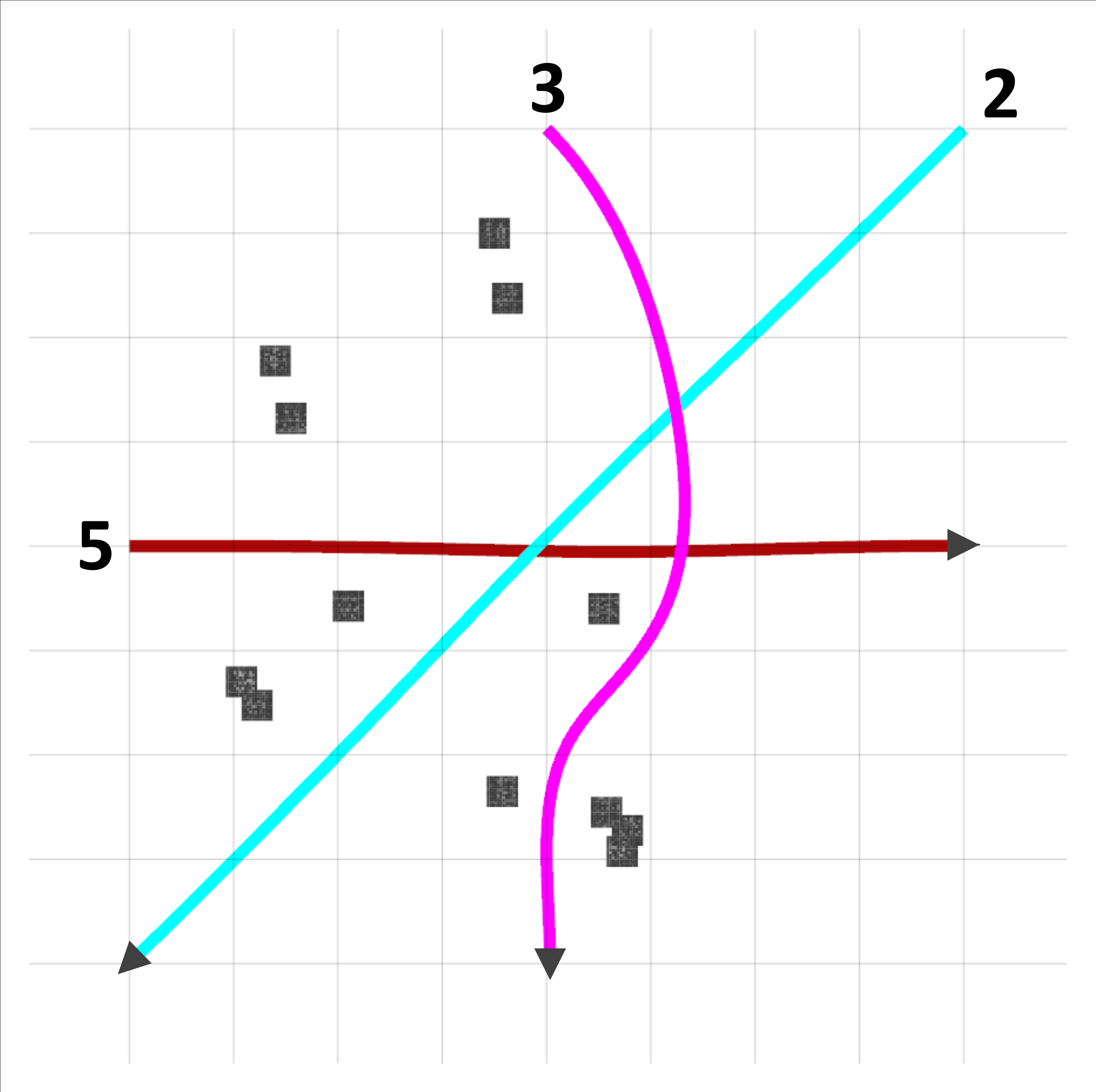}
		}
		\subfloat[]{
			\label{fig:sequential_2}
			\includegraphics[width=0.39\linewidth,trim={5 5 10 5},clip]{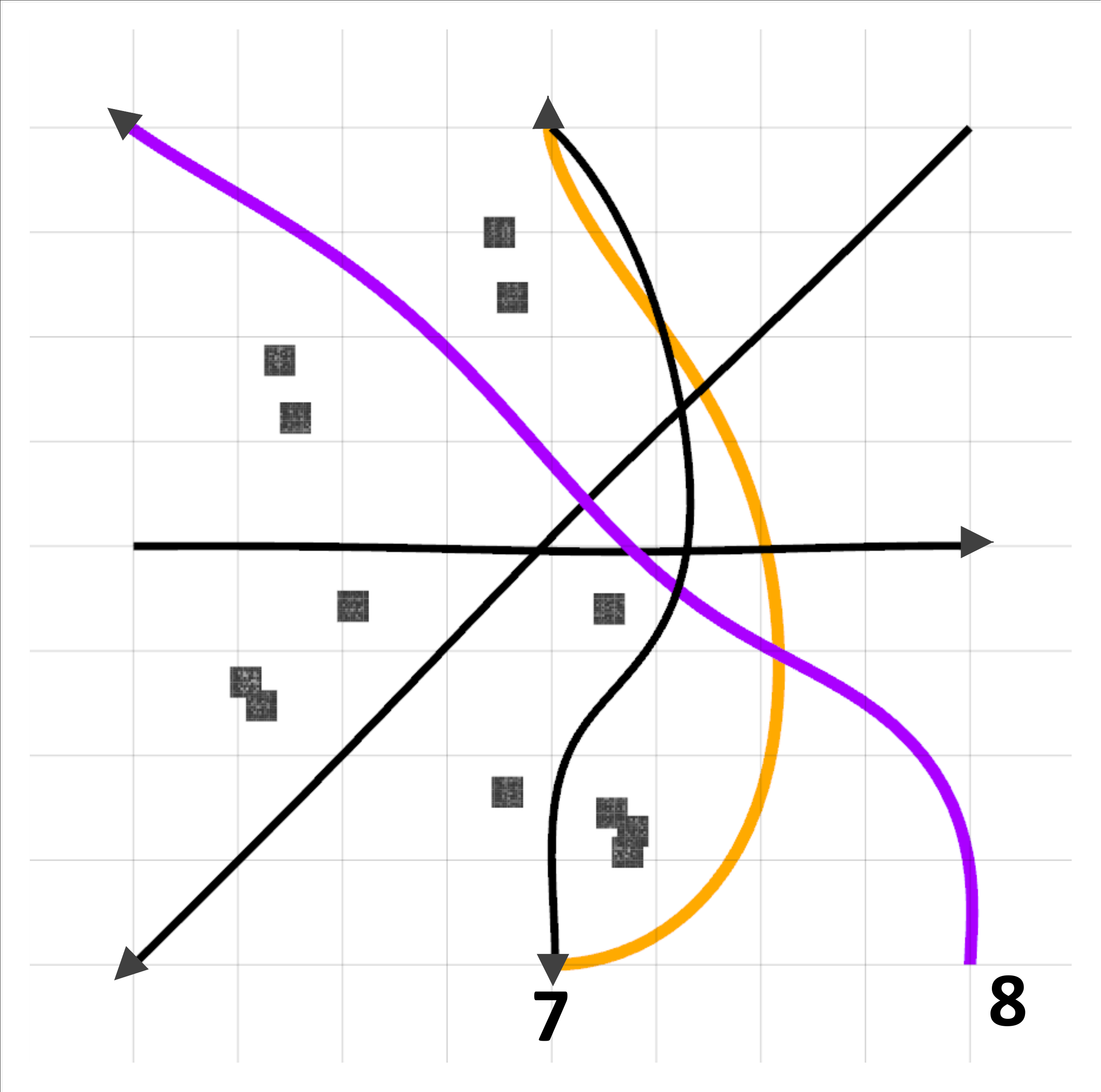}
		}
		\vspace{-0.41cm}
		\subfloat[]{
			\label{fig:sequential_3}
			\includegraphics[width=0.39\linewidth,trim={10 10 5 5},clip]{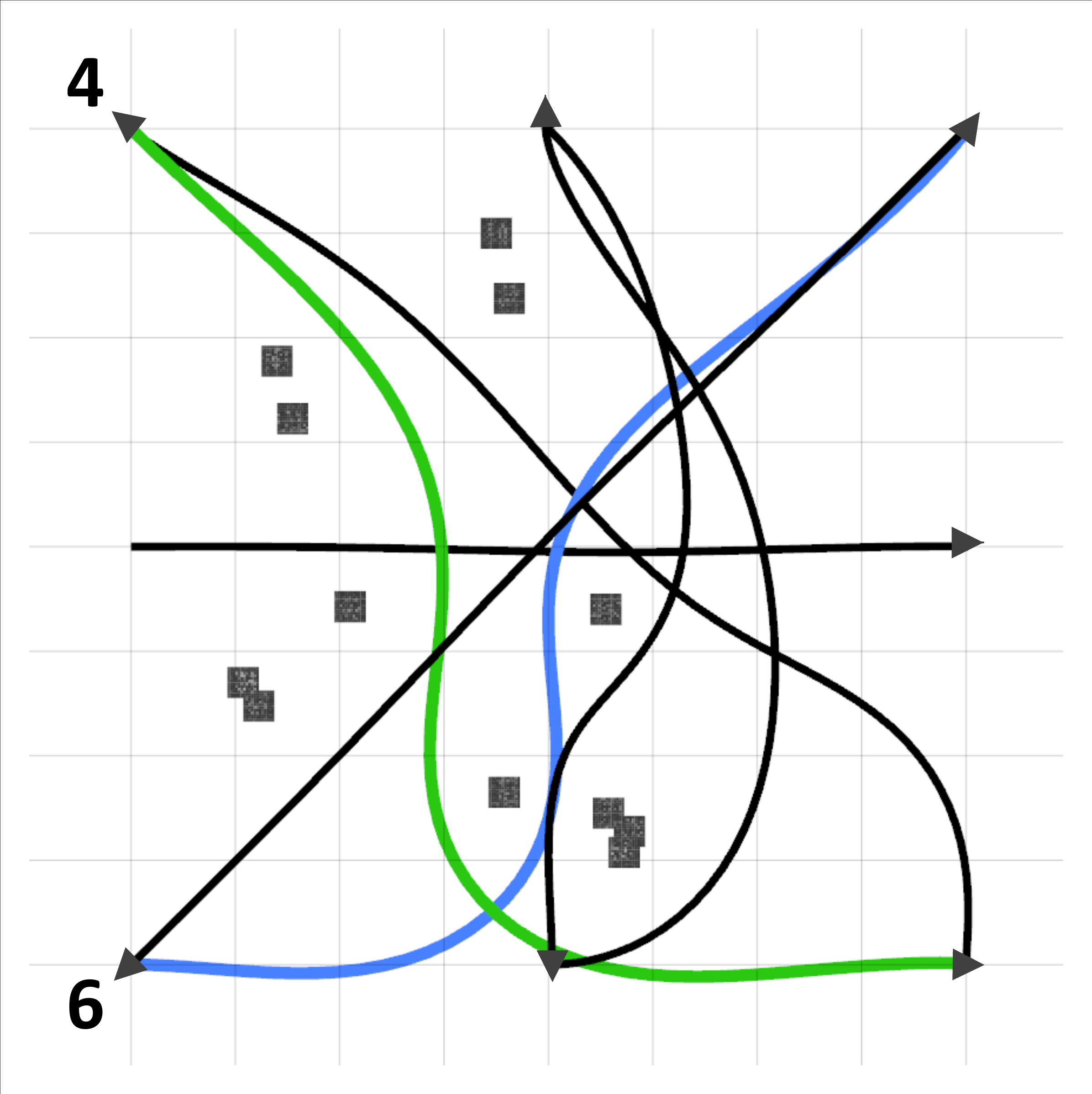}
		}
		\subfloat[]{
			\label{fig:sequential_4}
			\includegraphics[width=0.39\linewidth,trim={5 10 10 5},clip]{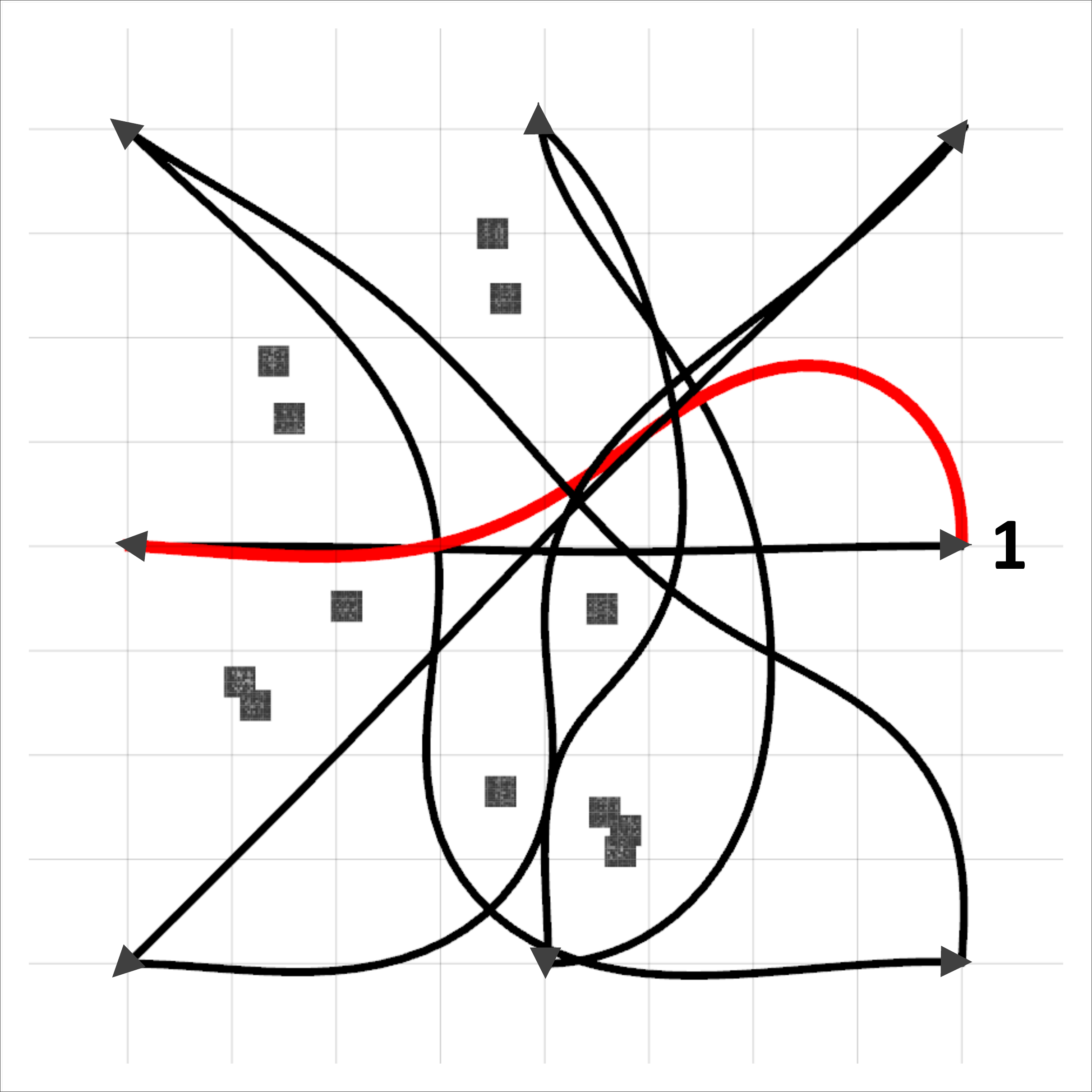}
		}
		\caption{An example of the prioritized trajectory optimization. In each iteration, the current group of robots need to avoid all higher-priority robots. The trajectories of the current group of robots and the higher-priority robots are depicted as colored lines and black lines, respectively. }
		\label{fig:example}
	\end{figure}
	
	As mentioned before, next we formulate and solve the problem \eqref{equ:MPC_new} for each group sequentially from high-priority to low-priority. Fig. \ref{fig:example} shows an example of the overall prioritized trajectory optimization process. Firstly, we group the 8 robots using Alg. \ref{alg:sequential_optimization}. The groups are listed from high-priority to low-priority, i.e. $[(2,3,5),(7,8),(4,6),(1)]$. In each iteration of the prioritized trajectory optimization, in addition to the inter-robot constraints inside the current group, inter-robot constraints between the current group of robots and the optimized higher-priority robots should also be considered to ensure no collisions. The optimized trajectories of each group of robots are shown in \crefrange{fig:sequential_1}{fig:sequential_4}.

	\section{Simulations}\label{simulation}
	
	\subsection{Implementation Details}\label{sec:simulation}
	The proposed algorithms are implemented in C++ and executed on a laptop running Ubuntu 16.04 with Intel i5-6300HQ @2.30GHz CPU and 12GB of RAM. We use OctoMap\cite{hornung2013octomap} to represent the occupancy map of the environment and an interior-point nonlinear programming solver IPOPT \cite{wachter2006implementation} to solve the trajectory optimization problem. Our code is released as an open-source package\footnote{\url{https://github.com/LIJUNCHENG001/multi_robot_traj_planner}}.

	In the simulation, the radius of the collision model of each robot is set to $R=0.15\text{m}$ and the velocities of each robot are limited by $v^{\text{max}}=1\text{m/s}$, $\omega^{\text{max}}=1\text{rad/s}$. We plan the discrete paths in the grid graph with grid size $D=1\text{m}$. In this case, the time interval $\Delta T$ is set to $1.6$s so that the condition \eqref{equ:feasiblility} is satisfied. Besides, in the safe corridor construction, each line segment is divided into $h=5$ equal segments.

	\subsection{Computational Efficiency and Solution Quality}
	
\begin{figure*}
		\centering
		\hspace{-0.55cm}
		\subfloat[$t=5$s]{
			\label{fig:t=5s}
			\includegraphics[width=0.27\linewidth,trim={25 24 25 24},clip,angle=90]{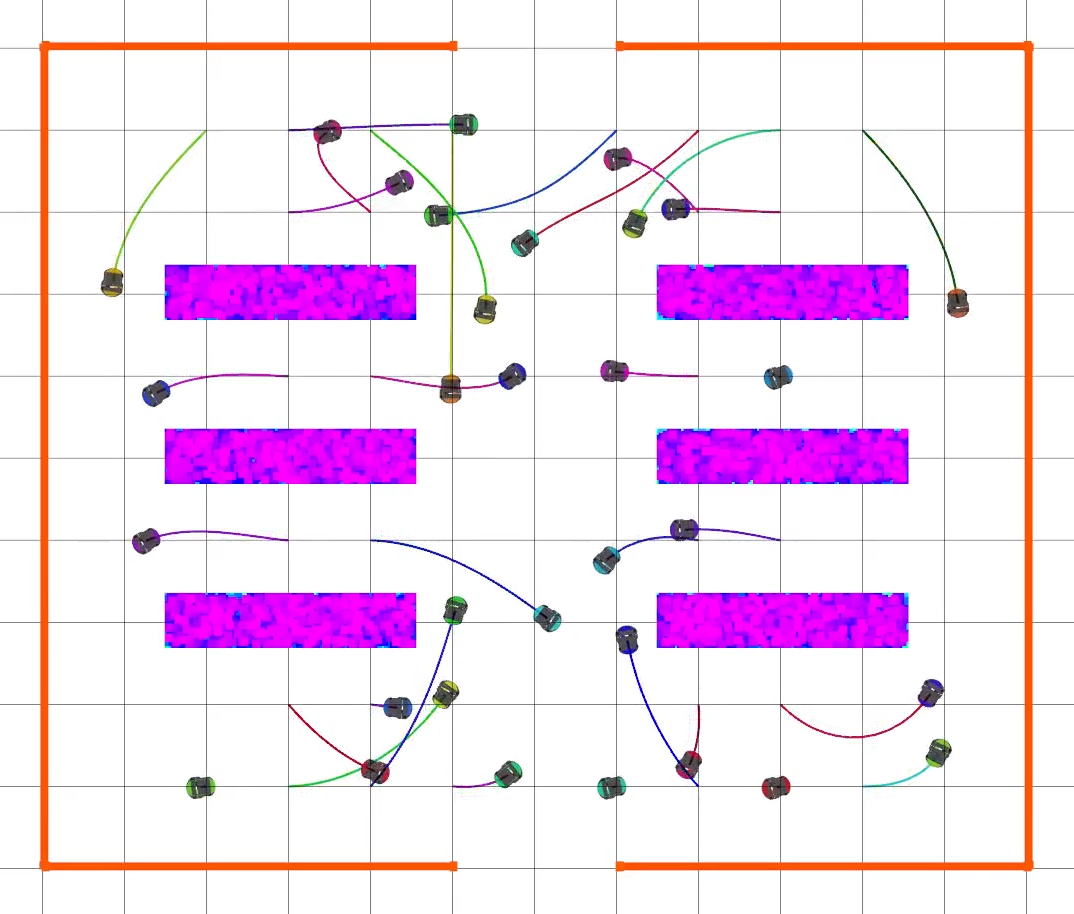}
		} \hspace{0.0cm}
		\subfloat[$t=10$s]{
			\label{fig:t=10s}
			\includegraphics[width=0.27\linewidth,trim={25 24 25 24},clip,angle=90]{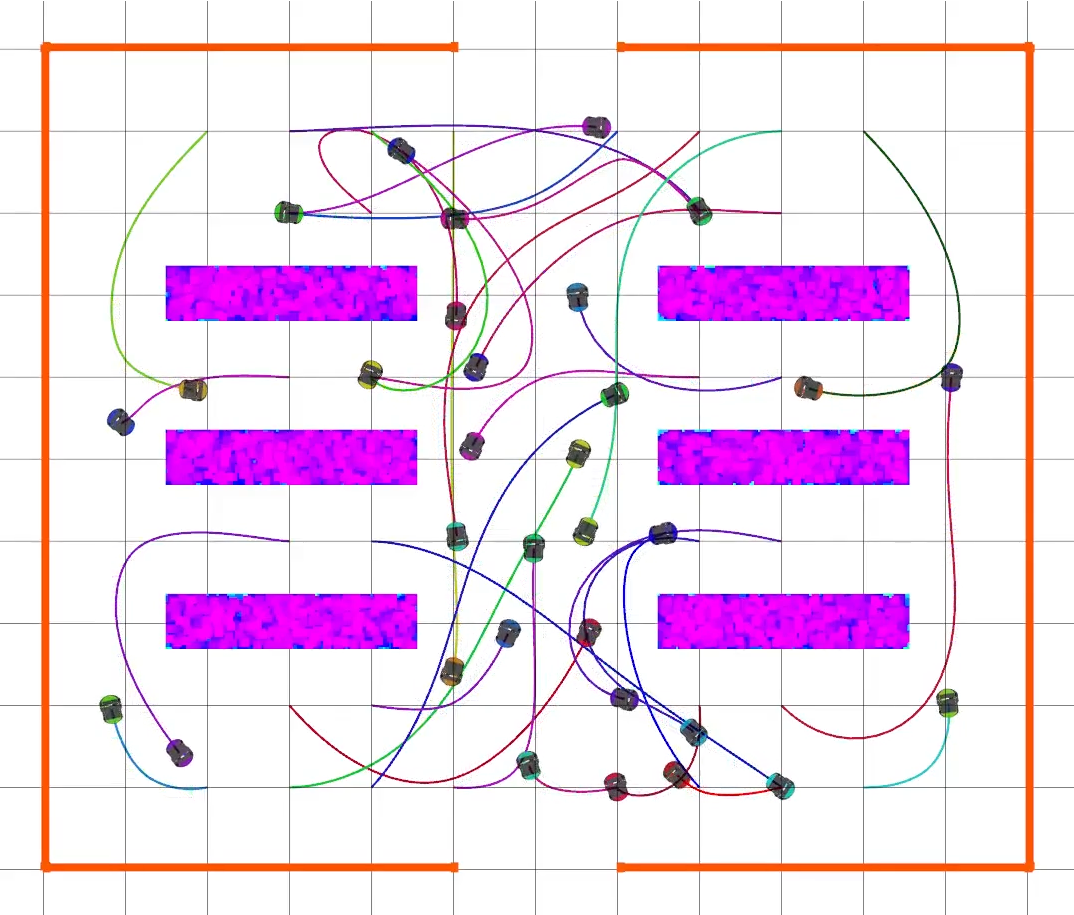}
		}\hspace{0.0cm}
		\subfloat[$t=15$s]{
			\label{fig:t=15s}
			\includegraphics[width=0.27\linewidth,trim={25 24 25 24},clip,angle=90]{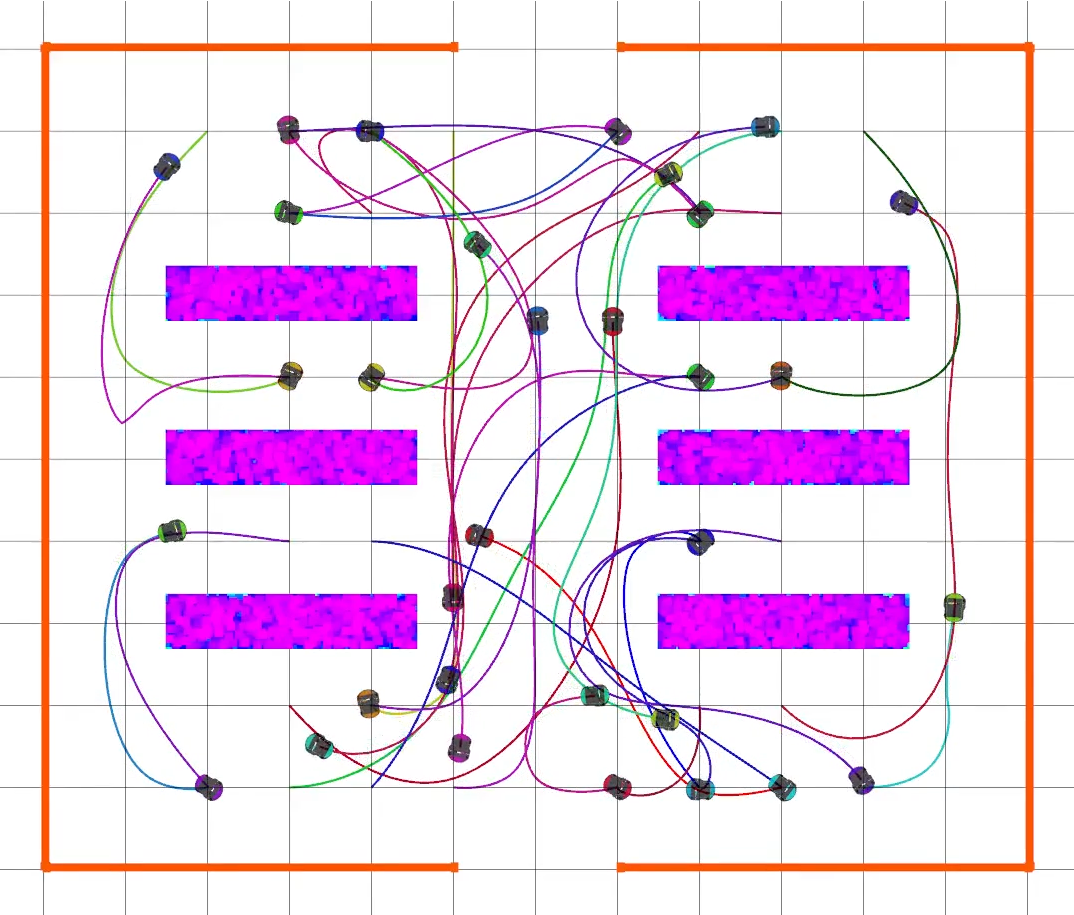}
		}\hspace{0.0cm}
		\subfloat[$t=20$s]{
			\label{fig:t=20s}
			\includegraphics[width=0.27\linewidth,trim={25 24 25 24},clip,angle=90]{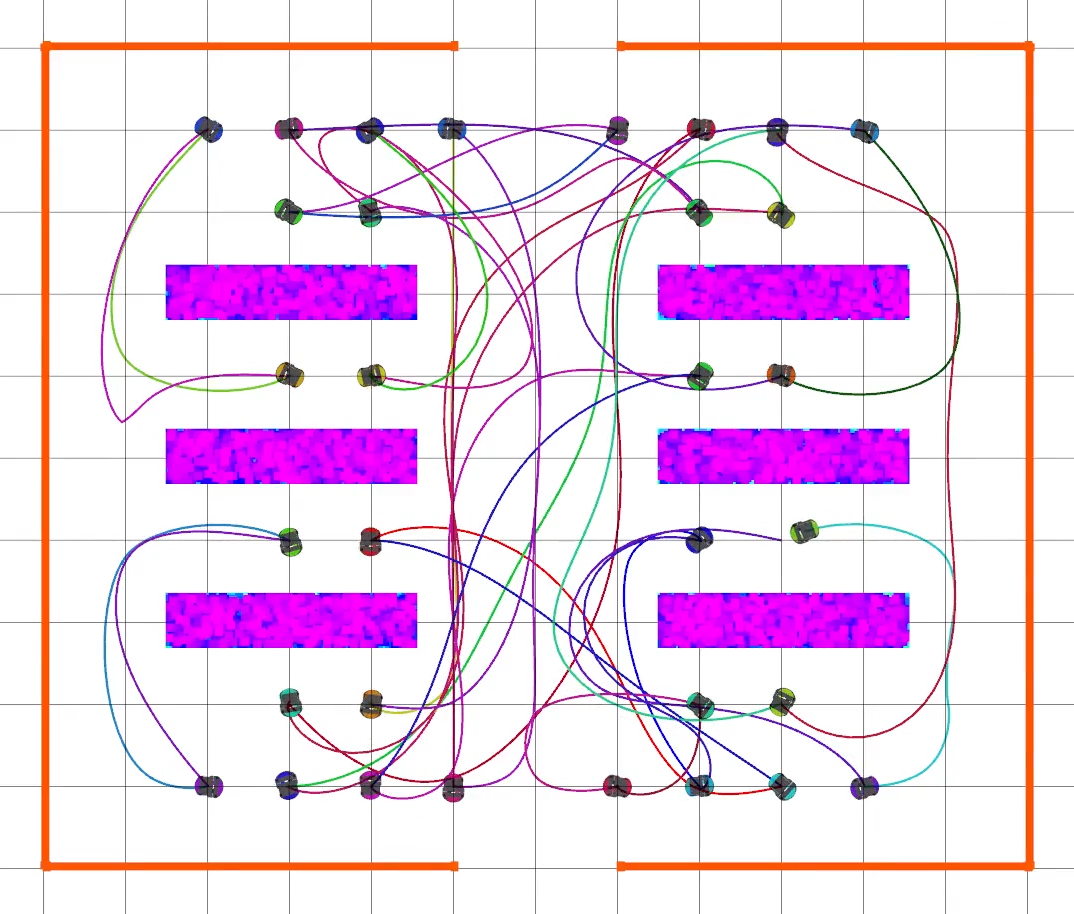}
		}\hspace{-0.3cm}
		\caption{Trajectories of 32 agents in a warehouse environment. }
		\label{fig:real_simulation}
	\end{figure*}

	Material handling in warehouses is the main target application of our developed approach. Here, we construct a warehouse environment to evaluate the performance of the proposed approach. The environment has a size of 10m $\times$ 12m and contains 6 shelves of size 3m $\times$ 0.6m. The start and goal positions of each robot are randomly assigned at the boundary of the environment or at the pick-up points near the shelves. Based on the environment settings, we set the suboptimal bound of ECBS as $1.5$ to fulfill the requirement on computation efficiency. Fig. \ref{fig:real_simulation} shows an example of 32 robots navigating in the simulation environment. We conduct the simulations for 40 times and calculate the average computation time, which is shown in Table \ref{tab:total_time}. The scalability of our approach is shown to be very good according to the results. The proposed approach takes about 6.1s to complete the whole trajectory planning process for 32 mobile robots.
	
	\begin{table}
		\newcommand{\tabincell}[2]{\begin{tabular}{@{}#1@{}}#2\end{tabular}}
		\caption{Computation time}
		\hspace{-0.25cm}
			\begin{tabular}{c||c|c|c||c}
				\hline \multicolumn{1}{c||}{\begin{tabular}[c]{@{}c@{}}Number \\ of agents\end{tabular}} & \multicolumn{1}{c|}{\begin{tabular}[c]{@{}c@{}}Discrete path \\ planning\;(s)\end{tabular}} & \multicolumn{1}{c|}{\begin{tabular}[c]{@{}c@{}}Safe corridor\\ construction\;(s)\end{tabular}} & \multicolumn{1}{c||}{\begin{tabular}[c]{@{}c@{}}Trajectory \\ optimization\;(s)\end{tabular}} & \multicolumn{1}{c}{\begin{tabular}[c]{@{}c@{}}Total \\(s)\end{tabular}} \\ \hline \hline
				4 & 0.001 & 0.008 & 0.142 & 0.151 \\
				8 & 0.002 & 0.019 & 0.360 & 0.381 \\ 
				16 & 0.006 & 0.031 & 1.042 & 1.079 \\ 
				24 & 0.326 & 0.046 & 2.526 & 2.898 \\ 
				32 & 2.175 & 0.062 & 3.853 & 6.090 \\
				\hline
		\end{tabular}
		\label{tab:total_time}
	\end{table}

	\begin{figure}
		\vspace{-0.63cm}
		\centering
		\subfloat[]{\hspace{-0.46cm}
			\label{fig:runtime}
			\includegraphics[width=0.57\linewidth,trim={120 80 90 80},clip]{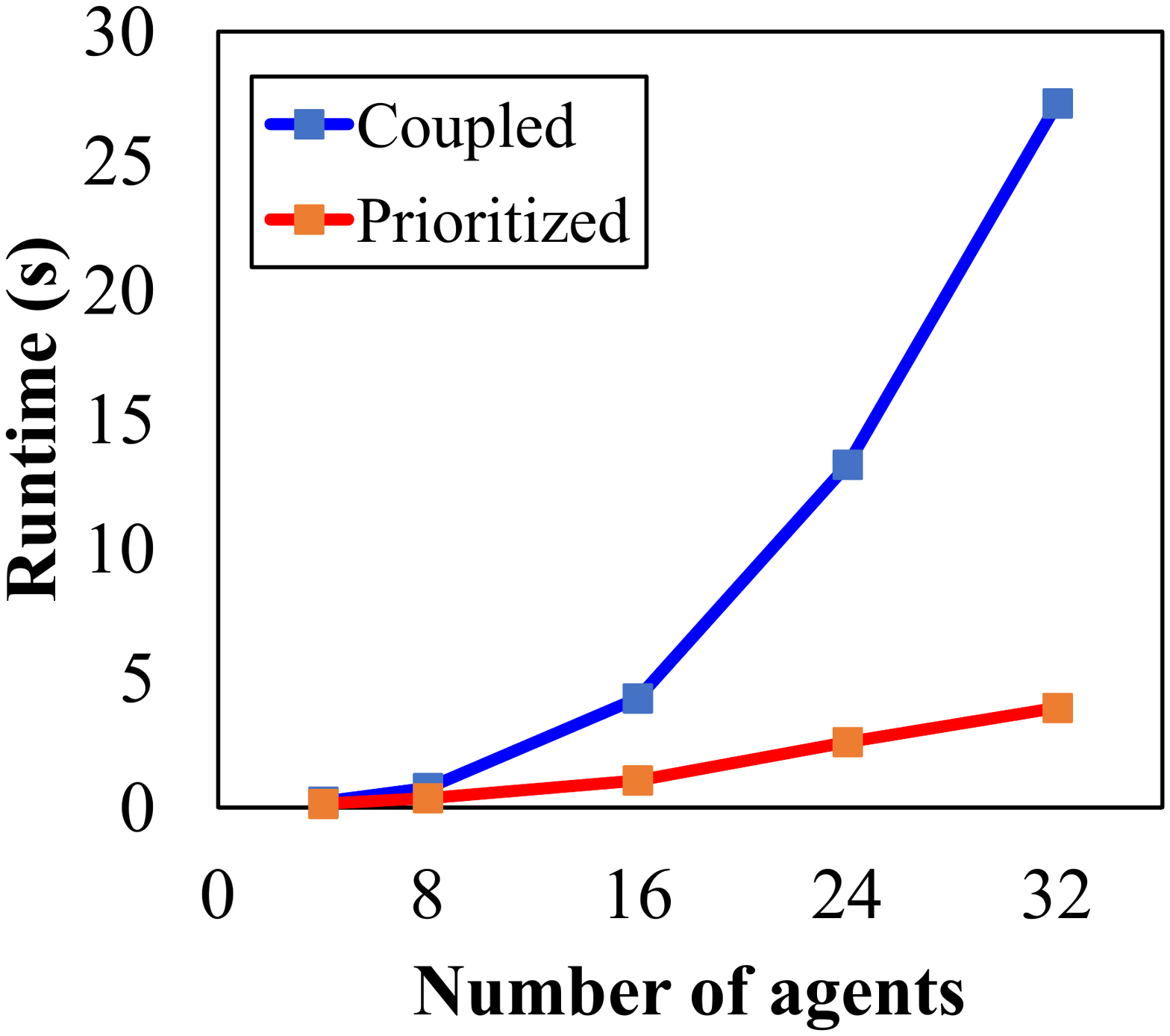}
		} 
		\subfloat[]{\hspace{-0.79cm}
			\label{fig:cost}
			\includegraphics[width=0.57\linewidth,trim={120 80 90 80},clip]{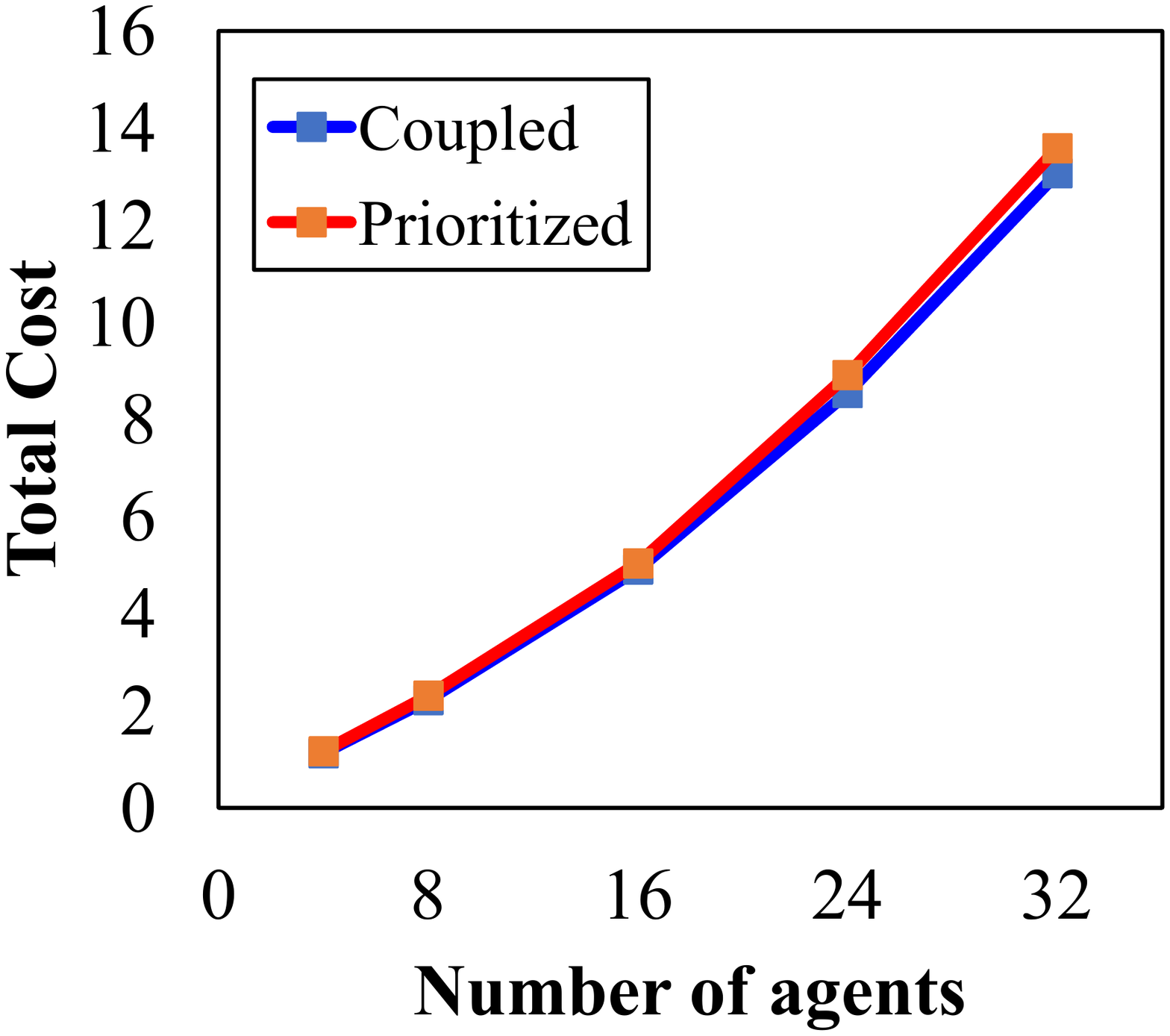}
		}
		\caption{Comparison of the proposed approach and the coupled trajectory optimization.}
		\label{fig:runtime_and_cost}
	\end{figure}
	
	In this paper, the method which directly solves the original large-scale trajectory optimization problem \eqref{equ:MPC_new} is referred to as coupled trajectory optimization method. As shown in Fig. \ref{fig:runtime}, the proposed prioritized optimization method shows much better computational efficiency compared to the coupled trajectory optimization method. Specifically, if the number of agents is 4 or less, the two methods achieve comparable performance. As the number of agents becomes larger, the coupling effect between the robots becomes stronger. Since the proposed method decouples the large-scale optimization problem, the computational efficiency is improved significantly. The coupled optimization takes about 27.2s to solve the 32-agents case while our method only takes about 3.9s. It can be observed that the runtime of the proposed method increases almost linearly. We mention that prioritized optimization uses a decoupled framework, so it typically leads to inferior solutions compared to the coupled optimization. As shown in Fig. \ref{fig:cost}, the total cost of the proposed method is on average 3.7\% higher than that of the coupled optimization, which is relatively small and acceptable in practice.

	\begin{figure}
		\centering
		\vspace{-0.1cm}
		\hspace{-0.5cm}
		\includegraphics[width=0.61\linewidth,trim={60 75 90 80},clip]{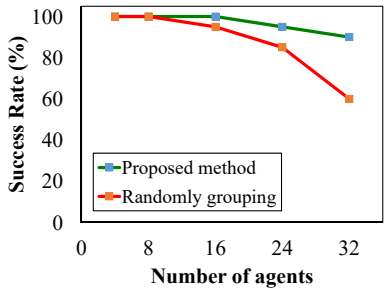}
		\caption{Success rate of the trajectory planning using different grouping strategies.}
		\label{fig:success_rate}
	\end{figure}

	\subsection{Success Rate}
	
	To avoid infeasible subproblems generated in the prioritized trajectory optimization, a novel grouping and priority assignment strategy is also developed in this paper. Fig. \ref{fig:success_rate} shows the success rate of the prioritized optimization using two different grouping strategies. Both methods can achieve 100\% success rate when the number of agents is small. As expected, as the number of agents increases from 8 to 32, the success rate of the method using randomized grouping decreases from 100\% to 60\%. However, because we assign priorities to the robots based on the specific scenario, the proposed method achieves higher success rates for all cases in the test environment. In the 32-agent case, the success rate of the proposed method is on average 90\%.

	\section{Experiment}\label{experiment}
	
	\begin{figure}
		\vspace{-0.15cm}
		\centering
		\includegraphics[width=0.8\linewidth,trim={350 220 380 200},clip]{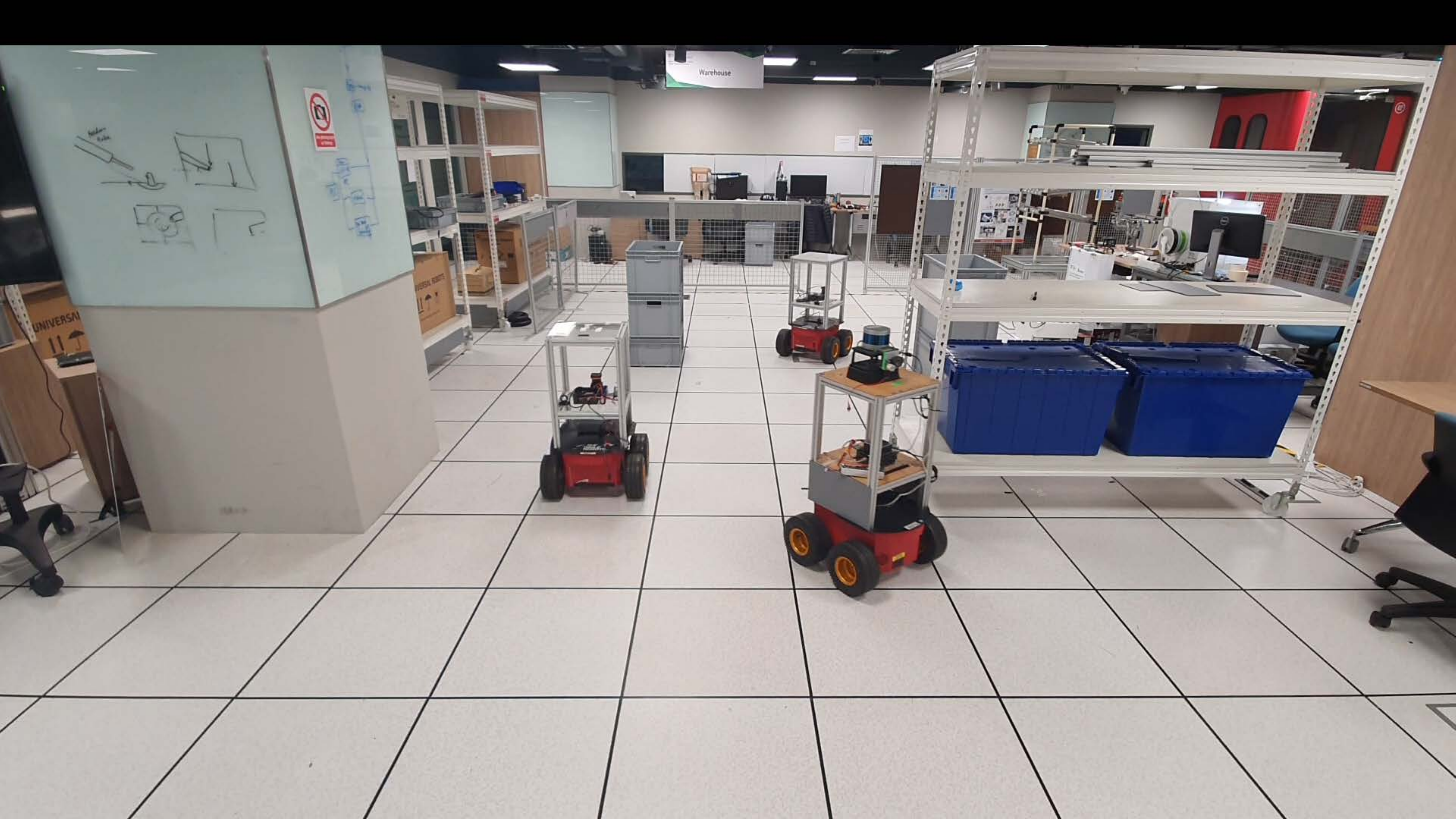}
		\caption{Three robots navigating in the real environment.}
		\label{fig:test_environment}
	\end{figure}

	\begin{figure}
		\vspace{-0.24cm}
		\centering
		\subfloat[]{\hspace{-0.23cm}
			\label{fig:planning}
			\includegraphics[width=0.48\linewidth,trim={733 30 294 115},clip]{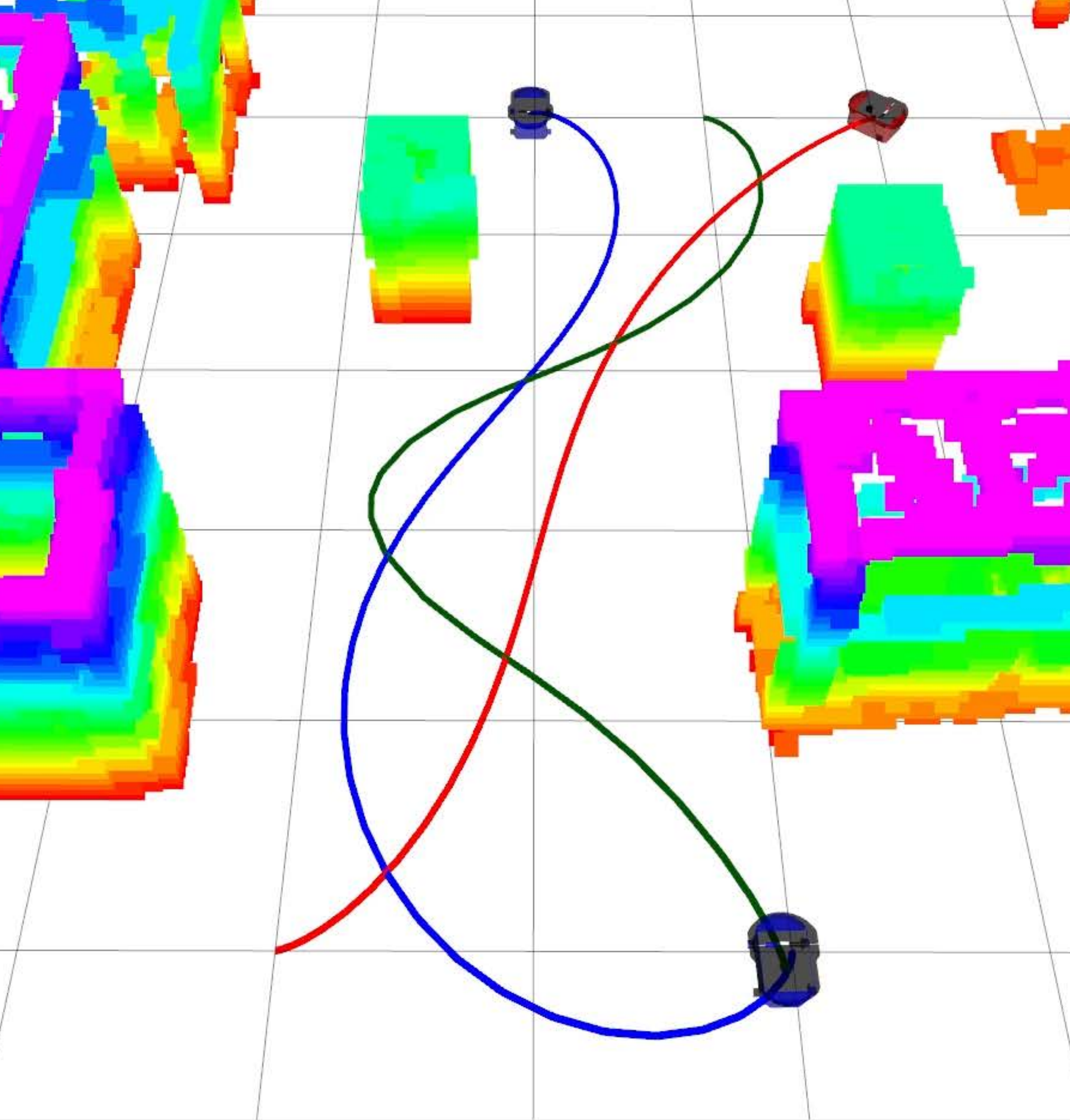}
		} 
		\subfloat[]{\hspace{-0.15cm}
			\label{fig:tracking}
			\includegraphics[width=0.488\linewidth,trim={0 0 0 0},clip]{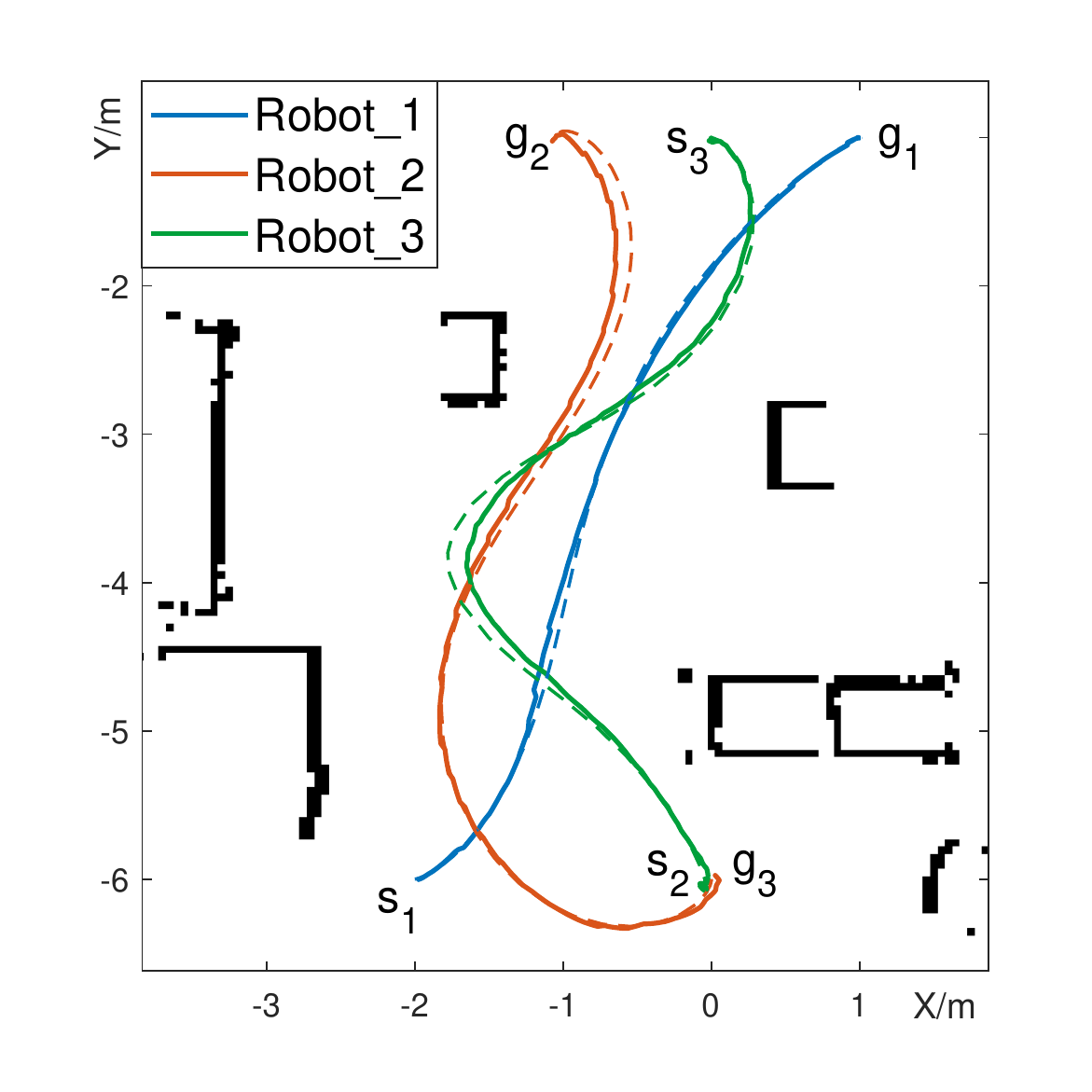}
		}
		\caption{Experiment results. Fig. 9a shows the 3D map of the environment and the planned trajectory. Fig. 9b shows the trajectory tracking results, where dash and solid lines represent the planned and real trajectories, respectively. }
		\label{fig:experiment}
	\end{figure}

	We conduct a real multi-robot navigation experiment with 3 Pioneer 3-AT robots in a 7m $\times$ 8m indoor testing area, which is shown in Fig. \ref{fig:test_environment}. One robot is equipped with a Velodyne VLP-16 3D lidar, and each of the other two robots is equipped with a Hokuyo 2D lidar which has a maximum range of 30m. At the beginning, the map of the test environment is constructed using a 3D lidar odometry and mapping method, LeGO-LOAM \cite{shan2018lego}. We construct the 3D map because obstacles of different heights need to be considered in the trajectory planning algorithm to guarantee safety. The radius of the collision model of each robot is set to $R=0.3$m and the velocities of each robot are limited by $v^{\text{max}}=0.6$m/s, $\omega^{\text{max}}=0.6$rad/s. Besides, the time interval $\Delta T$ is set to $2.65$s in the experiment and all the other parameters remain the same as described in Sec \ref{sec:simulation}. The planned trajectories in the experiment are shown in Fig. \ref{fig:planning}. We upload the trajectories to the three robots and use a MPC-based trajectory tracking control method \cite{li2019mpc} to execute the trajectories. The localization of the robots is obtained through AMCL ROS package. As shown in Fig. \ref{fig:tracking}, the team of robots can track their reference trajectories accurately and complete the task without any collision. The experimental video is available at \url{https://youtu.be/GRl3LM8xBUQ}.

	\section{Conclusion}\label{conclusion}
	In this paper, we presented an efficient trajectory planning algorithm for multiple non-holonomic robots navigation in obstacle-rich environments. The trajectory planning problem is decoupled as a front-end path searching and a back-end nonlinear trajectory optimization. We adopt a multi-agent path searching method to find collision-free time-optimal initial paths, which are further refined into smooth and dynamically feasible trajectories. A prioritized trajectory optimization method is proposed to improve the scalability of the back-end algorithm. We split the team of robots using a novel grouping and priority assignment strategy, and then solve the optimization problem sequentially. The effectiveness and superiority of the proposed method are validated in both simulations and real-world experiments.
	
	In future work, we plan to integrate our work with distributed multi-agent navigation methods, so that the trajectory of each robot can be replanned online to handle unknown dynamic obstacles in the environment.

	
	
	
	
	\balance
	
	\bibliographystyle{IEEEtran}	
	\bibliography{ugv}

\end{document}